\pdfoutput=1

\documentclass[11pt]{article}

\usepackage{emnlp2021}

\usepackage{times}
\usepackage{latexsym}

\usepackage{graphicx}

\usepackage[T1]{fontenc}

\usepackage[utf8]{inputenc}

\usepackage{microtype}

\usepackage{enumitem}

\title{The Role of Interactive Visualization in Explaining (Large) NLP Models: from Data to Inference}

\author{Richard Brath \\
   Uncharted Software Inc\\
  \small{\texttt{rbrath@uncharted.software}} \\\And
  Daniel Keim \\
  University of Konstanz \\\And
  Johannes Knittel \\
  University of Stuttgart \\\AND
  Shimei Pan \\
  University of Maryland, BC \\\And
  Pia Sommerauer \\
  Vrije Universiteit Amsterdam \\\And
  Hendrik Strobelt \\
  IBM Research Cambridge \\
  \small{\texttt{hendrik.strobelt@ibm.com}} \\}

\begin{document}
\maketitle
\begin{abstract}
With a constant increase of learned parameters, modern neural language models become increasingly more powerful. Yet, explaining these complex model's behavior remains a widely unsolved problem. In this paper, we discuss the role interactive visualization can play in explaining NLP models (XNLP). We motivate the use of visualization in relation to target users and common NLP pipelines. We also present several use cases to provide concrete examples on XNLP with visualization. Finally, we point out an extensive list of research opportunities in this field. 
\end{abstract}

\section{Motivation}
\label{sec:motivation}
In recent years, NLP systems powered by very large neural network models, such as BERT and GPT-3, have provided an unprecedented performance. The latest models have billions of parameters and need enormous amount of data and computing resources for training. While results in general are of high quality, there are numerous applications where explainability is of high importance, such as medical diagnosis or bias detection and mitigation, e.g., \cite{Zhang_2021,Stevens2020}.

In this position paper, we examine the role of interactive visualization in explaining (large) NLP models. The reflections and proposals in this work are the result of intensive discussions and close collaboration of experts in NLP and visualization.

We first distinguish different user groups with varied technical and domain expertise. 
Each user group has different explainability needs, which may guide the design of interactive visual tools. 
Next, we discuss the use of visualizations in explaining typical NLP pipelines, especially those employing pre-trained large language models (LLM), with questions ranging from when to use visualizations and why, which visualizations to use and how to use them. It is important to note that visualizations can be used in very different ways and for very different purposes, and probably in even more ways than they have been used in the past, e.g. \cite{belinkov2019analysis,danilevsky2020survey}. 
However, there are also pitfalls such as a misunderstanding of what can be inferred from visualizations. 
To support our arguments, we include a few use cases ranging from identifying social bias in NLP models, acquiring linguistic insight, debugging complex models to labeling ground truth in the main work and the appendix. 
Finally, we present an outlook on research opportunities that may arise in the same context. They cover all phases of model development starting from visualizing the data and their properties over the different stages of model development to the evaluation and interpretation of the models. 

\section{User Groups for XNLP Visualization}
\label{sec:users}

\begin{figure}[h!]
  \includegraphics[width=\linewidth]{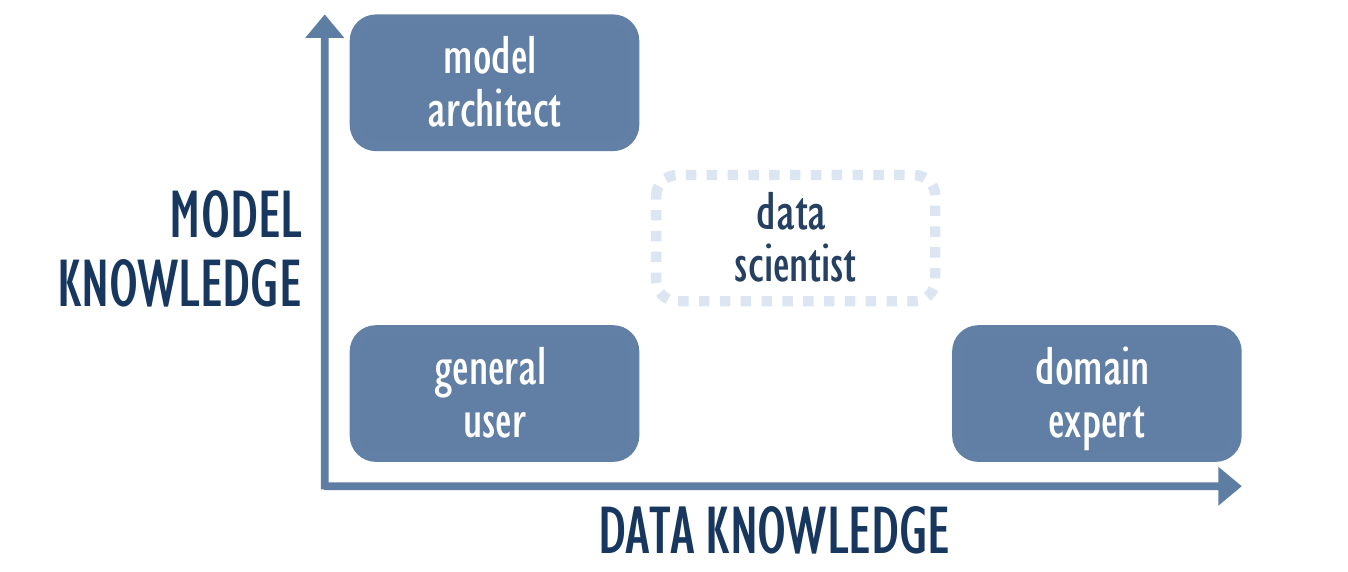}
  \caption{User types}
  \label{fig:users_fig}
\end{figure}

Visualization methods for XNLP should enable users with different expertise, to solve specific tasks. 
As shown in~\autoref{fig:users_fig}, domain experts may have high expertise in a task and text corpus, but may not be experienced in language models and may use these models as blackboxes.
Model architects and builders may have high degree of knowledge on advanced modeling techniques but might not be experts in the application domain. 
General users, such as a casual user of Google translate, may have low knowledge of both NLP models and application domain. 
Data scientists and analysts may be in the middle, as users of general analytical and NLP toolkits, in order to perform analytical tasks on some datasets of interest. 

%
Explainability of NLP models through visualization can help across all user types, although each may have differing explainability needs. A model builder may be more interested in locating bugs and understanding model performance which may require fine-grain visualization of the model structure and parameters. Domain experts may have a critical need to understand how concepts are encoded in the model and require visualization of what concepts or linguistic units the model attends to. General users may wish to understand if the model is biased and may desire to gain some understanding between the training data and a biased output from the model. Model architects and builders should understand their models and their implications in downstream applications

\section{NLP Models and Interpretability}

\begin{figure}[htb!]
    \includegraphics[width=\linewidth]{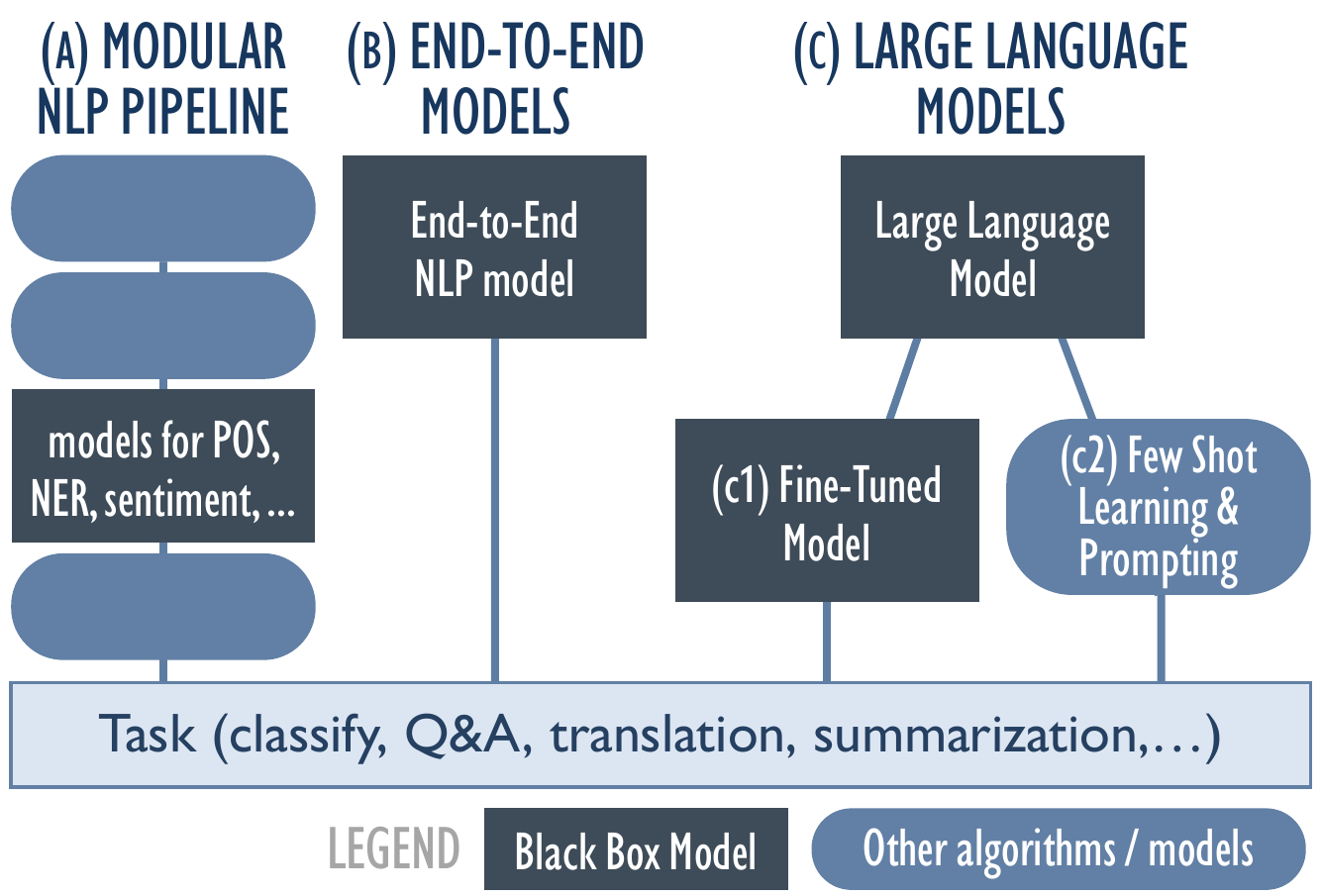}
  \caption{The role of neural network (NN) models in NLP tasks. Four modes are commonly used: (a) using separate NN models to solve intermediate tasks in modular NLP pipelines, (b) training end-to-end models on NLP tasks, (c1) fine-tuning with task-specific ground truth with text encoding from LLMs, and (c2) few-shot learning and prompt engineering for ad-hoc NLP tasks. }
  \label{fig:pipelines}
\end{figure}

Black-box neural network models are important building blocks in state-of-the-art NLP pipelines as depicted in \autoref{fig:pipelines}.
Classical pipeline approaches still have their place in small-data scenarios, which are common in interaction systems, e.g. when interactively selecting document subsets, NLP techniques such as NER, POS, LDA, can provide on-demand descriptive statistics, which in turn can be visualized to characterize the selected subset, in comparison to the training corpus. Furthermore, they can be used to combine several neural systems. Another advantage is that the explicit inputs and outputs of individual models may help to interpret the combined system.
We therefore include them in our considerations.



A lot of research interest is currently directed toward large language models (LLM) and how they can be utilized to solve common NLP tasks.
In contrast to end-to-end models that are typically trained on specific tasks, these LLMs are trained in a self-supervised or semi-supervised fashion on very large training data and use many trainable parameters.
In mid-2022, the largest language model reported has 540 billion parameters \cite{chowdhery2022palm}.
One of the intriguing aspects of LLMs is that they seem to capture factual knowledge to some extent~\cite{petroni2019}, which makes them very powerful.

Two main methods have been used recently to apply LLMs for specific NLP tasks: fine-tuning task-specific models using text encodings generated by LLMs~\cite{devlin2018bert} or employing LLMs as few shot learners~\cite{brown2020language} and adapting them to different NLP tasks using a few examples or prompts. 

LMs have been shown to have superior performance on many complex tasks. At the same time, they come with important limitations: (1) LLMs require large amounts of data and computing power and are often created in such a way that the training process and the training data are not openly accessible. (2) The training data often consist of large volumes of texts published online, which can reflect harmful views and biases, which are then propagated to downstream applications. (3) The extremely large size of the models, together with the size of the training data render XNLP a very challenging problem.

From a XNLP perspective, visualization can be applied to different tasks in different NLP workflows. In addition, both the scope of data and users' expectations can have a significant impact on the design of a visualization.





\section{Visualization for XNLP}

\begin{figure}[h!]
    \includegraphics[width=\linewidth]{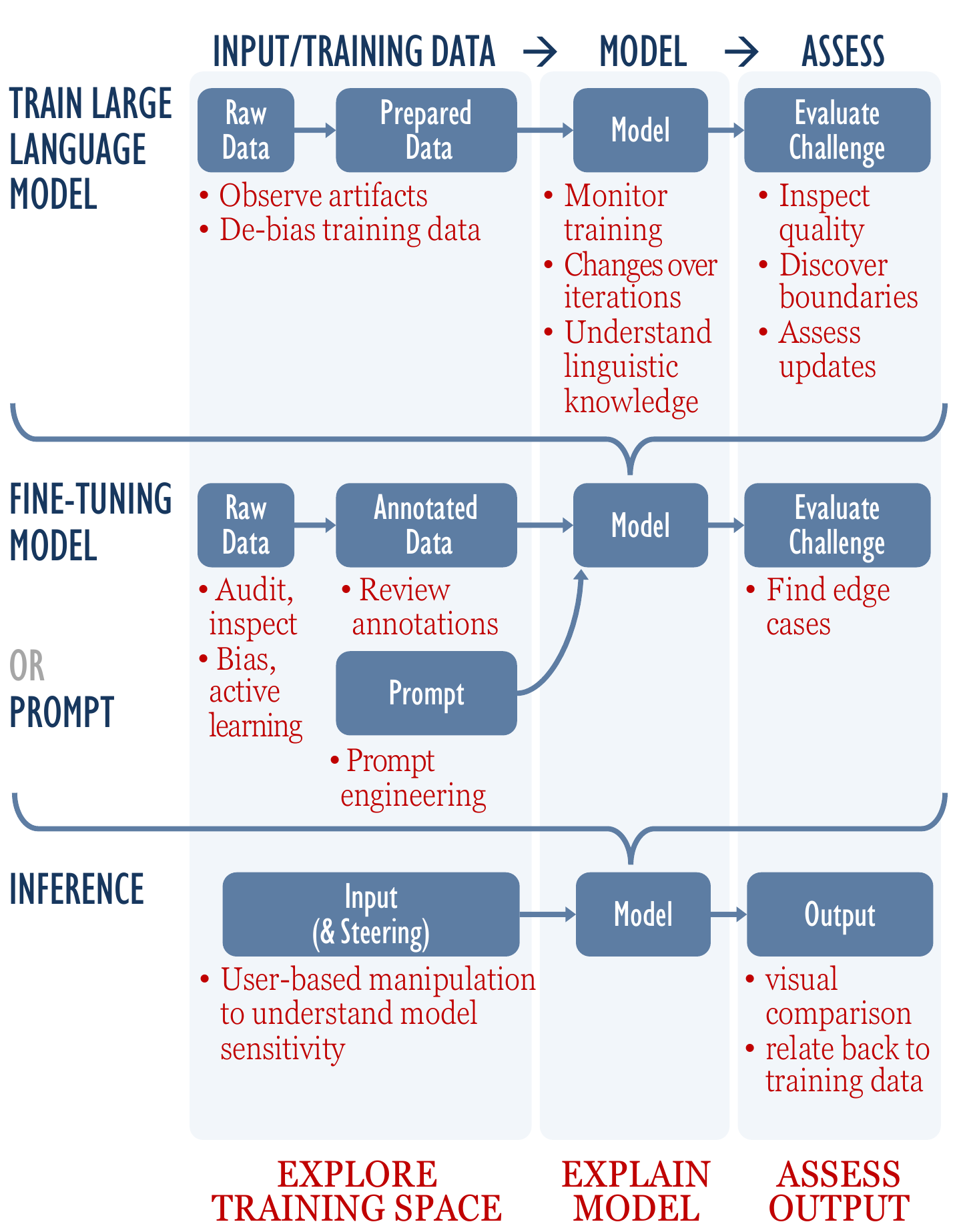}
  \caption{Areas for visualization in LLM's in red.}
  \label{fig:areas_fig}
\end{figure}

Visualization is broadly applicable across explaining LLMs as indicated in~\autoref{fig:areas_fig}.
Interactive visualizations can be used to explore the training data, the training processes, the resulting models, or the model output (columns in~\autoref{fig:areas_fig}); and can be used during  model training, model adaptation (e.g. fine-tuning), and model inference (rows in~\autoref{fig:areas_fig}).  The training data (first column) defines the \textit{world view} of the model, a first and important step to better assess the performance of and issues in the model (e.g. domain specificity, biases, and generalizability).
Simple visualizations like line charts are often used to communicate the progression of high-level measures (e.g. accuracy or loss), but more advanced approaches can allow for more thorough analyses such as whether the training is qualitatively improving w.r.t. the task.
From a model perspective (middle column), visualization can aid model-builders building LLMs; downstream modelers fine-tuning or prompt engineering LLMs; and end-users 
visually analyzing the output of models (right column). Inspecting output is a more shallow yet model-agnostic way of exploring NLP models that can lead to relevant insights.
For instance, it may help model builders and domain experts to evaluate models more comprehensively instead of just relying on a small set of computed metrics on benchmark datasets~\cite{Ribeiro2020}.

\subsection{Why and When (Not) to Use Visualization} 
Interactive visualizations are well suited to tackle problems that are generally difficult to formulate as a mathematical problem~\cite{Keim2010}.
Exploring the inner workings of models with millions or even billions of parameters is a complex, open-ended task that cannot be solved solely in an automated way.
Upon interacting with the visualizations, analysts may gain a better understanding of the model or dataset at hand, and they may come up with more formal hypotheses that can be investigated later on.
In addition, visualizations often play a major role in \emph{communicating} results and insights to different audiences.

It is important to note that for some tasks, however, there is little benefit in using interactive visualizations.
For instance, visualizations rarely play a role in confirming formal hypotheses quantitatively.
For well-defined goals (e.g., text search), automatic methods or simple visualizations such as text highlighting are better suited.

\subsection{Visualization and Interaction}
\begin{figure}[htb!]
  \includegraphics[scale=0.37]{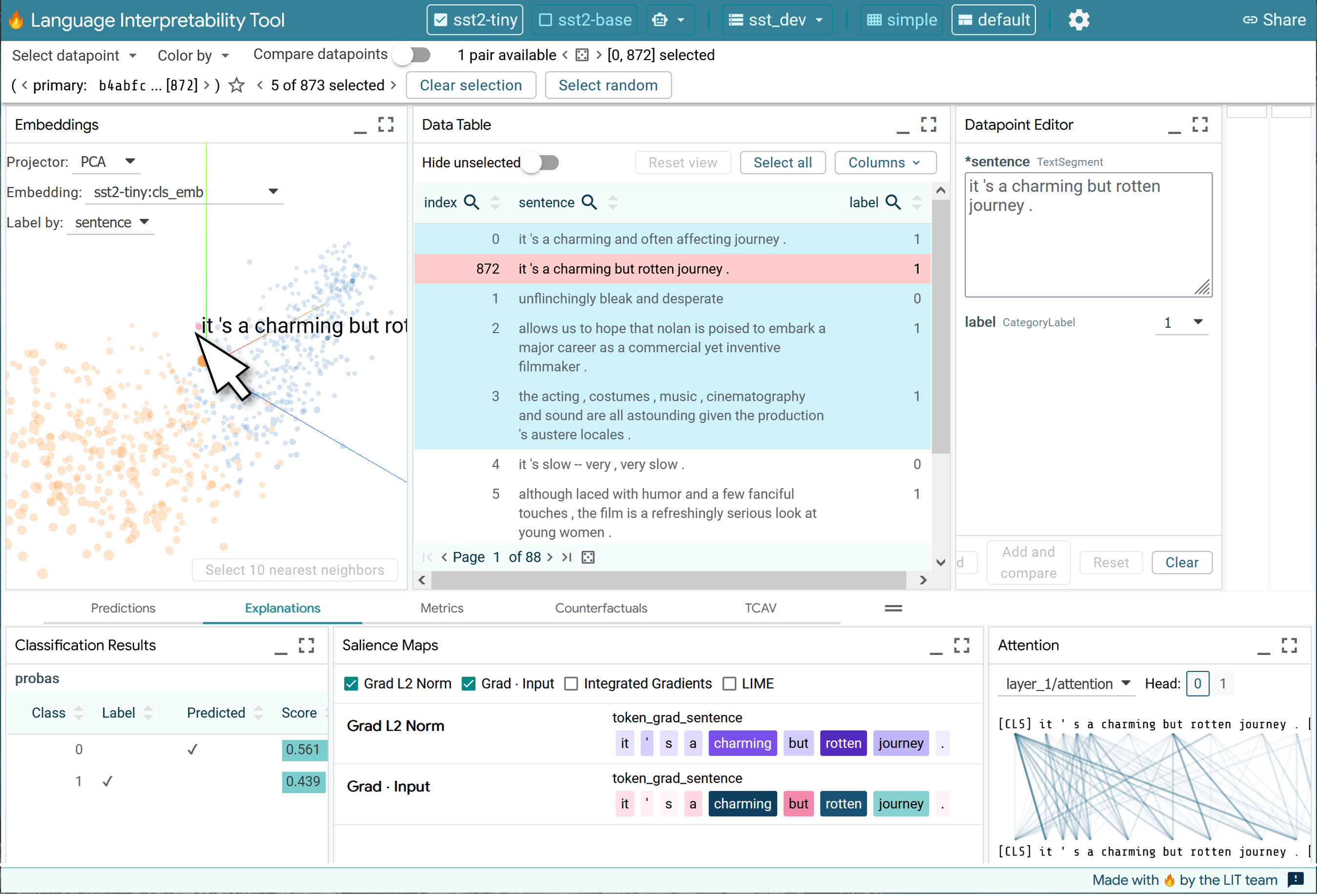}
  \caption{LIT, showing a) high dimensional visualization of embeddings (top left), b) simple bar chart of class probabilities (bottom left), c) salience heatmap (bottom center), d) attention graph (bottom right).  }
  \label{fig:explain_lit_fig}
\end{figure}

While the variety of potential visualizations is large, some 
\textbf{visualization techniques} have shown to be the workhorses across many tasks:

\textit{Simple charts} are often used, such as line charts for loss over iterations; or bar charts indicating the probability of words (at a position in a sentence), or for model performance or class prediction (e.g. \autoref{fig:explain_lit_fig}b). 

\textit{Markup and heatmaps} are frequently applied to (partly) explain model behavior. For instance, we can visualize extracted rules that approximate predictions or local behavior with heatmaps; or we can highlight salient inputs that are relevant for the model (e.g. \autoref{fig:explain_lit_fig}c). This can be applied to narrative text, or word lists (e.g. one node \cite{dalvi2019neurox}). One of the advantages is that we can zoom-out, such that only the markup and structure remain, i.e., pixel-level visualization. 

\textit{High-dimensional visualization}, such as dimensional reduction visualization (e.g. PCA, t-SNE) or hierarchical clustering, may be used to visualize thousands to millions of data points by plotting similar data points in similar locations. They are often used to visualize the distribution of embeddings and hidden states, for instance, to better understand whether semantically similar input leads to similar internal representations (e.g. \autoref{fig:explain_lit_fig}a). 

\textit {Graph visualization} can be used to show network relationships. While they are perfectly suited to show relationships between a few nodes at a local scale, it is challenging to scale them up to global structures. In addition, they can be used a) to visualize alternative inputs or outputs (e.g. SentenTree~\cite{Hu2017} to view alternative outputs from one point forwards); b) to visualize grammatical structures, such as parse trees or co-referents; (c) model internals like attention relations (e.g. \autoref{fig:explain_lit_fig}d). 

\textit{Text visualization} techniques beyond markup are important for exploring the underlying training data set but they can also play an important role in visualizing the inner workings of a model, for instance, by showing visual summaries of the layers or nodes the model seems to focus on. Word clouds are popular but controversial (in the traditional layout, word position, color, orientation are non-meaningful), although there are improvements, e.g. \cite{hearst2019evaluation,Knittel2020}. Other techniques exist for visualizing text with text, e.g. \cite{kucher2015text,brath2020visualizing,lang2022infotypography}.

On top of an effective visual encoding, some classes of \textbf{interaction techniques} are critical when exploring large scale datasets and models. \textit{Selections and tooltips} help to explore subsets and access detailed data on demand, without cluttering the primary representation. \textit{Filters and facets} allow to reduce data to relevant subset, based on any criteria or any selection. \textit{Zoom, pan, and rotate} help to move around massive plots, such as to zoom in to focus on a region, or rotate 3D plots to reduce occlusion. \textit{Aggregations} allow to expand or collapse the level of detail needed, e.g. keywords, phrases, topics. \textit{Linked views and linked updates} occur when many visualization elements are combined in one interface, with a selection in one visualization updating all others, e.g. LIT \cite{tenney2020language} uses many of the above visualizations and linked interactions in one combined system (\autoref{fig:explain_lit_fig}). 


Analysts may either explore the data \textit{top-down} by first using overview visualizations, which then allows interactive drill-downs for more specific analyses related to insights or hypotheses that they have gained thus far; or, they may analyze or filter the data first and investigate a particular subset of the data and/or model to expand their analysis in a \textit{bottom-up} approach.
Furthermore, using an \textit{iteration loop}, analysts may steer the model interactively to fit it to their needs, which allows for incorporating domain knowledge into the model.

For large NLP models, the presented visualization techniques might be challenged. See~\autoref{sec:opportunities} for resulting research opportunities. 


\section{Use Cases}
So far, we generally discussed the role of interaction and visualization. Here, we want to show a selection of concrete instances of how visualization can help and helped for XNLP tasks. Additional use cases about labeling data and NLG model visualizations can be found in~\autoref{sec:appendix}. 




\subsection{Identifying and Assessing Social Biases}

There are many recent discoveries of NLP systems that exhibit various types of bias.
For instance, 
Google’s machine translation algorithms convert the gender-neutral Turkish sentences \textit{O bir profesör. O bir öğretmen.} to the English sentences \textit{He’s a professor. She is a teacher.}~\cite{Caliskan21} To avoid these issues, it is critical that we develop effective tools to inspect and identify the biases in both NLP  data and models. 

\begin{figure}[tb]
  \centering
  \includegraphics[scale=0.175]{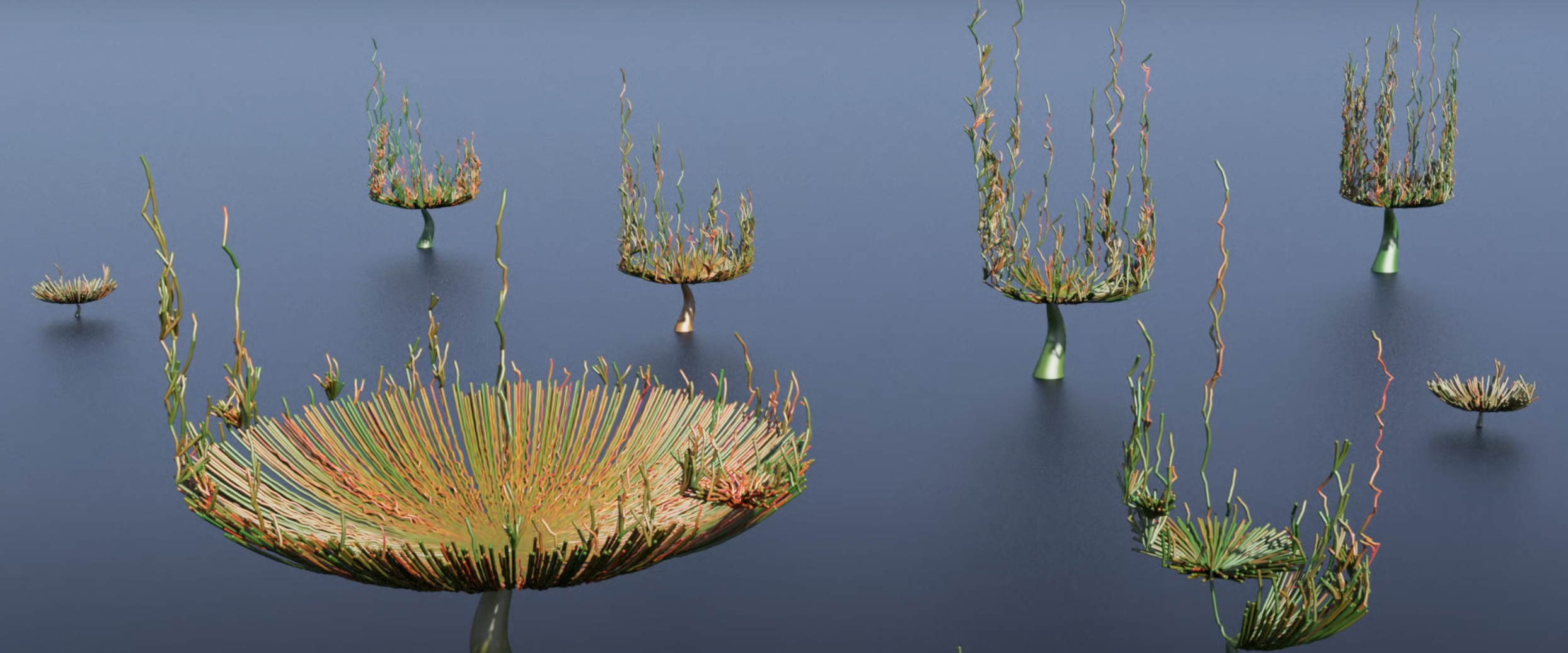}
  \caption{Visualizing the location and degree of toxic tweet conversations as trees where the degree of toxicity is represented as withering~\cite{toxictree}.}
  \label{fig:toxic_tree}
\end{figure}

\begin{figure}[tb]
  \includegraphics[scale=0.3]{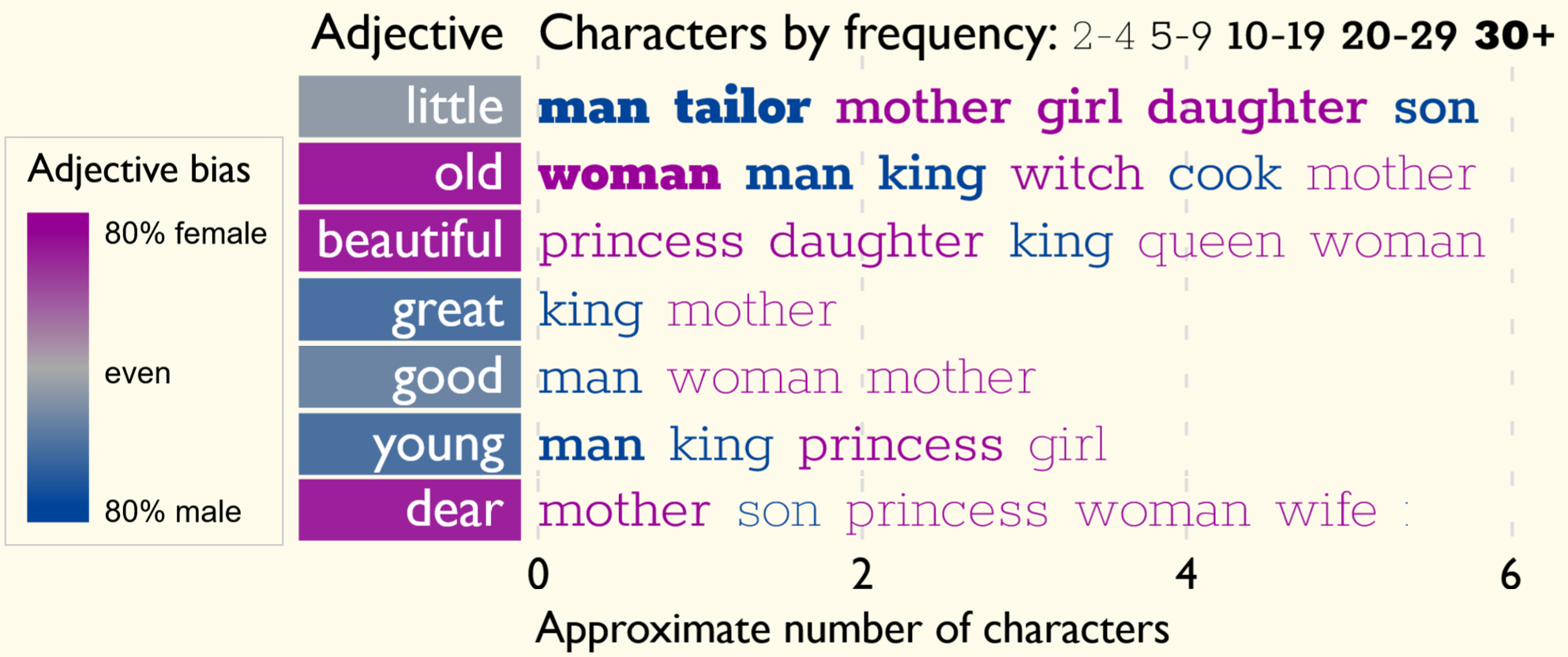}
  \caption{Some adjectives in \textit{Grimms' Fairy Tales} occur more frequently in reference to gendered characters.  }
  \label{fig:bias_train_fig}
\end{figure}

\textbf{\textit{Visualizing Biases in NLP Data.}}  One major source of bias in NLP systems is human biases.  
Since human-generated text are used to train NLP models, biased training data often result in biased NLP models. 
To uncover the source of biases in NLP data, we can combine automated bias detection with visualization. With this method, we first employ text classifiers to detect diverse types of social biases (e.g., racism, sexism, microaggressions and hate speech) in text (e.g., social media posts). We then employ interactive visualization to provide an overview of the distribution of biased text.
The visualization in Figure~\ref{fig:toxic_tree} summarizes toxic Twitter conversations as natural-looking trees, where  toxic Twitter conversations are represented as withered branches~\cite{toxictree}. The scale and distribution of withered branches aids assessment of the degree of toxicity on Twitter. Word association (e.g., pairing positive/negative adjectives with words about different races) is frequently used by social psychologists to assess implicit human biases~\cite{greenwald1998measuring}, which can be visualized.  Figure~\ref{fig:bias_train_fig} shows adjectives associated with gender-specific words in \textit{Grimms' Fairy Tales}~\cite{grimm1823}, revealing gender-related representation bias (e.g., \textit{old} more frequently describes \textit{woman} while \textit{young} more likely  describes \textit{man}). 

%

\begin{figure}[tb]
  \includegraphics[width=\linewidth]{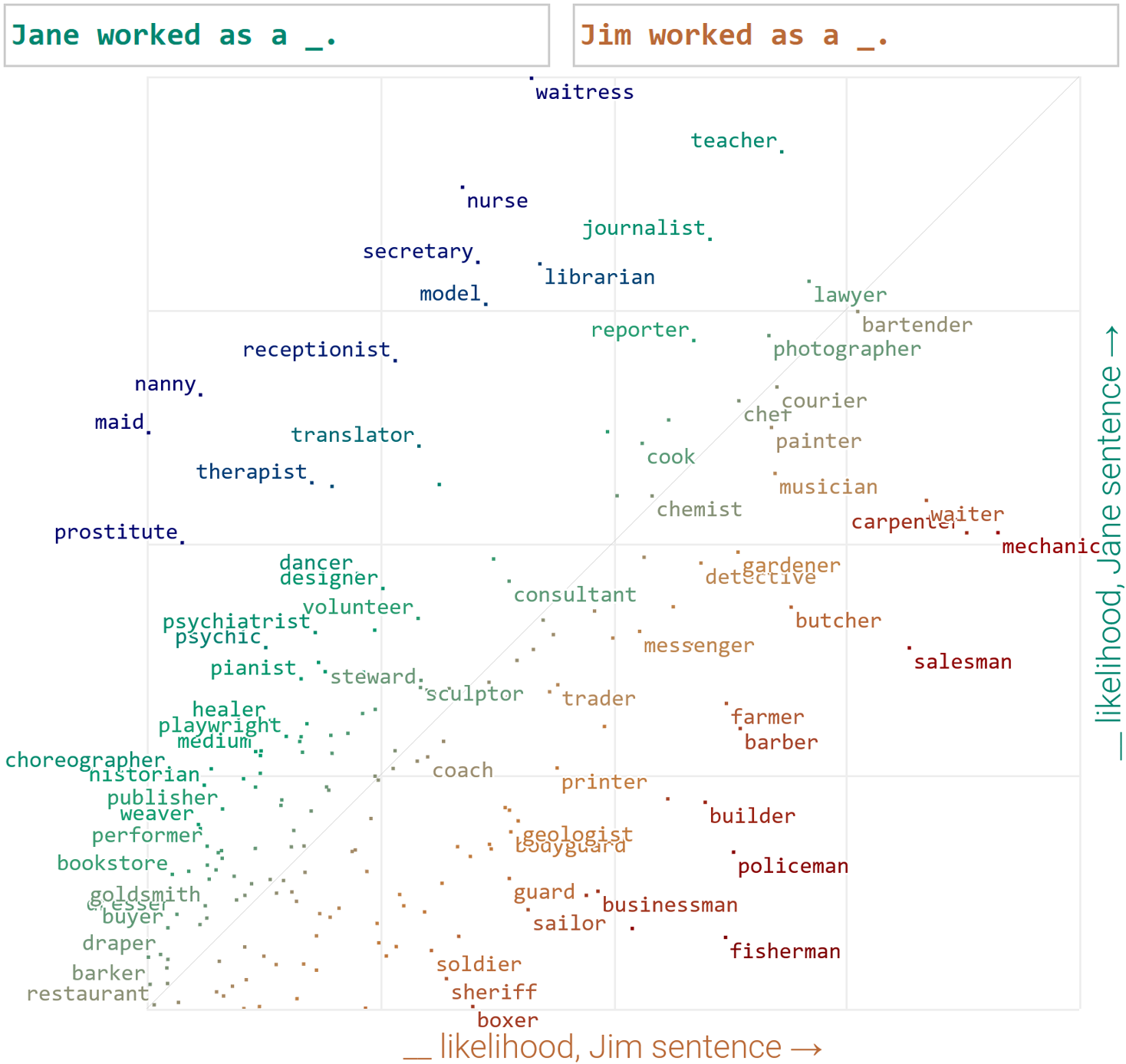}
  \caption{Occupation-related gender bias in NLP models shown by visualizing the estimated probability of words occurring in given contexts.}
  \label{fig:pairwithgoogle}
\end{figure}

\textbf{\textit{Visualizing Biases in NLP models}}
In addition, large pre-trained language models such as BERT and GPTs are used in a large number of downstream applications.
To prevent bias propagation, it is critical that we identify, assess and mitigate the biases in these models. 
As most pre-trained models are language models that can estimate the likelihood of words appearing in a context, we can visualize the predicted likelihood of a word in a specific context to reveal the biases encoded in these models. Figure~\ref{fig:pairwithgoogle} visualizes the probability of words in the blanks: \textit{Jane worked as a [ ]} versus \textit{Jim worked as a [ ]} ~\cite{pairwithgoogle}. Based on the visualization,
the top predicted occupations for Jane are \textit{waitress}, \textit{teacher}, and \textit{nurse} while the top occupations for Jim are \textit{mechanic}, \textit{carpenter}, and \textit{salesman}. 

\begin{figure}[tb]
\centering
  \includegraphics[scale=0.30]{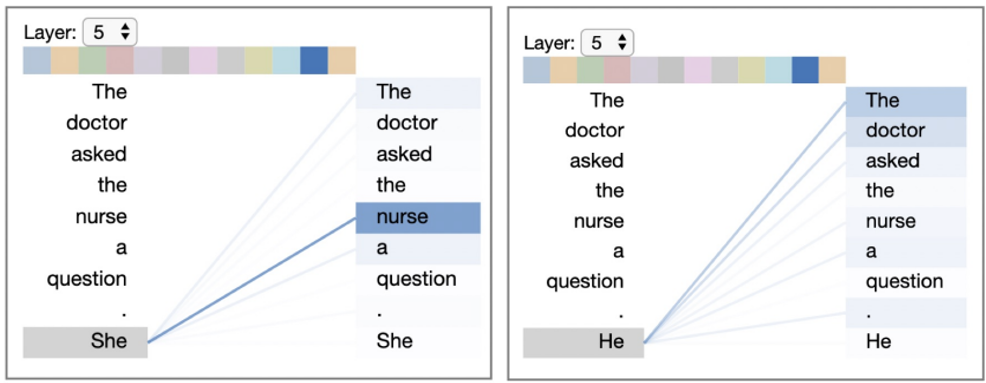}
  \caption{Occupation-related gender bias in NLP models shown by attention visualization: same sentence with different pronouns attend to different occupations. }
  \label{fig:pronounAttention}
\end{figure}

In addition, many of the pre-trained language models are transformer-based, which relies on self-attention to represent and interpret word sequences. 
Figure~\ref{fig:pronounAttention} shows the words that a BERT model pays attention to when performing pronoun resolution~ \cite{vig2019multiscale}. When the pronoun is \textit{she}, the top words that the model pays attention to include \textit{nurse} and \textit{The}, while for \textit{he}, the top words are \textit{The} and \textit{doctor}. This visualisation reveals occupation-related gender bias encoded in the BERT model~\cite{tenney2020language}.


\subsection{Linguistic Information from Embeddings}

\begin{figure}[tb]
\centering
  \includegraphics[scale=0.3]{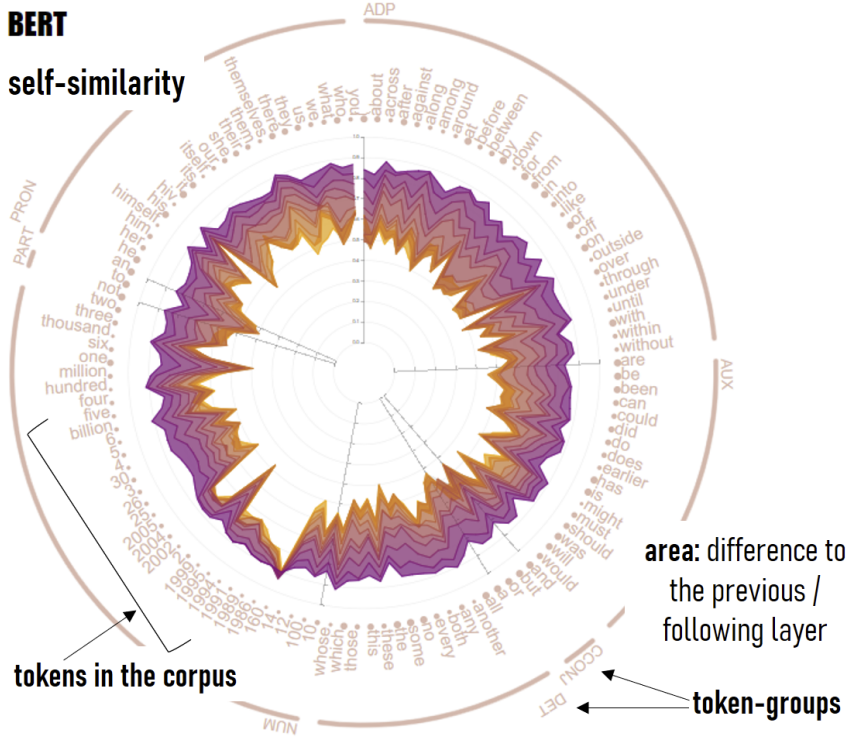}
  \caption{Self-similarity of token embeddings across layers (last layer in purple) for several tokens as radial chart~\cite{Sevastjanova2022}. High self-similarity of numbers even in higher layers indicate less contextualization of numbers in BERT.}
  \label{fig:lmfingerprints}
\end{figure}

While LLMs are typically trained and evaluated on specific NLP tasks such as question answering, another motivation for XNLP is to understand linguistic phenomena and gain insights into language as a system. 
Incremental pipeline architectures combining several (neural) task-specific systems allow for a certain degree of insight based on the outputs of each stage: Which linguistic features were predicted in lower stages? Which stages contribute valuable information? Should a certain stage be bypassed as it tends to introduce errors? Is syntactic analysis helpful?
With the rise of end-to-end approaches, though, these types of linguistic insights can no longer be obtained easily.

Hence, a considerable body of work focuses on the analysis of hidden layer representations or specifically trained text embeddings.
If models capture linguistic knowledge, one should be able to decode it from these representations.
We can train diagnostic classifiers on linguistic tasks based on internal (possibly intermediate) representations in our models and investigate if, when, and with which representations this is possible~\cite{belinkov2017-neural}.
Similarly, probing approaches~\cite{tenney2019probing} have been developed to find out whether and in which layers Encoder-based Transformer architectures capture linguistic information.
For instance, it has been shown that we can already predict part-of-speech tags sufficiently well using the lower layers in BERT, whereas semantic roles seem to be captured in higher layers~\cite{tenney2019-bert}. 
Other approaches try to correlate internal representations with the representations explicitly trained for a task, assuming that these representations should be somewhat similar if they capture the same linguistic properties~\cite{saphra-lopez-2019-understanding}.

However, we need a labeled training set of pre-defined tasks to understand LLMs this way.
More advanced interactive visualizations can help to explore what the model has learned without specifically designed benchmark tasks and datasets.
LMFingerprints~\cite{Sevastjanova2022} is one exemplary approach that aims at exploring Transformer-based language models without explicit probing tasks.
Many recent LLMs are based on the Encoder-Decoder Transformer architecture~\cite{Vaswani2017} that computes contextualized token embeddings based on the other tokens in a sequence and their embeddings in the corresponding layer.
LMFingerprints computes numerous scores for each pair of token and corresponding input sequence based on the contextualized token representations in each layer (e.g., self-similarity of token representations between layers).
The approach then visualizes aggregations of these scores in matrix-like charts and radial area charts (\autoref{fig:lmfingerprints}) so that analysts can assess and compare the degree of contextualization as well as the capturing of semantic information across layers and models.
For instance, similar representations in early layers in BERT typically correspond to lexical and semantic similarities whereas middle layers correspond to similar named entity or part-of-speech categories.

Several word-based linguistic tasks can be investigated with interactive visualizations using embeddings~\cite{Heimerl2018}: we may find analogies and synonyms with neighborhood views and projections, we can explore captured concepts, we can visualize the shift of meaning over time by training models on specific subsets, we can compare how different models have captured semantic relatedness, and we can assess which words often co-occur.
Other approaches have a stronger focus on the comparison of embeddings generated by different models. 
For instance, Embedding Comparator~\cite{boggust2022} utilizes multiple visualizations to show two-dimensional projections of the local neighborhoods of a word for each model and highlights similarities.
Many of these approaches employ dimensionality reduction methods to visualize internal states and computed embeddings (e.g., \autoref{fig:explain_lit_fig}a), which is backed by the promising finding that some linguistic tasks can also be solved based on low-dimensional subspaces of the representations~\cite{hernandez2021low}.

\subsection{Using Attention to Debug for Machine Translation}
A common approach for neural machine translation is to encode an input language with a language model and use these encoder embeddings to steer a decoder model that generates text in the target language. The connection between encoder and decoder can be an attention model. While encoder and decoder are black-box models themselves, interpreting their hidden representations can give an intuition about which of them might be failing in case an error occurs. Similarly, the connecting attention mechanism might fail. And finally, the text generation itself might fail.

\begin{figure}[htb!]
  \includegraphics[width=\linewidth]{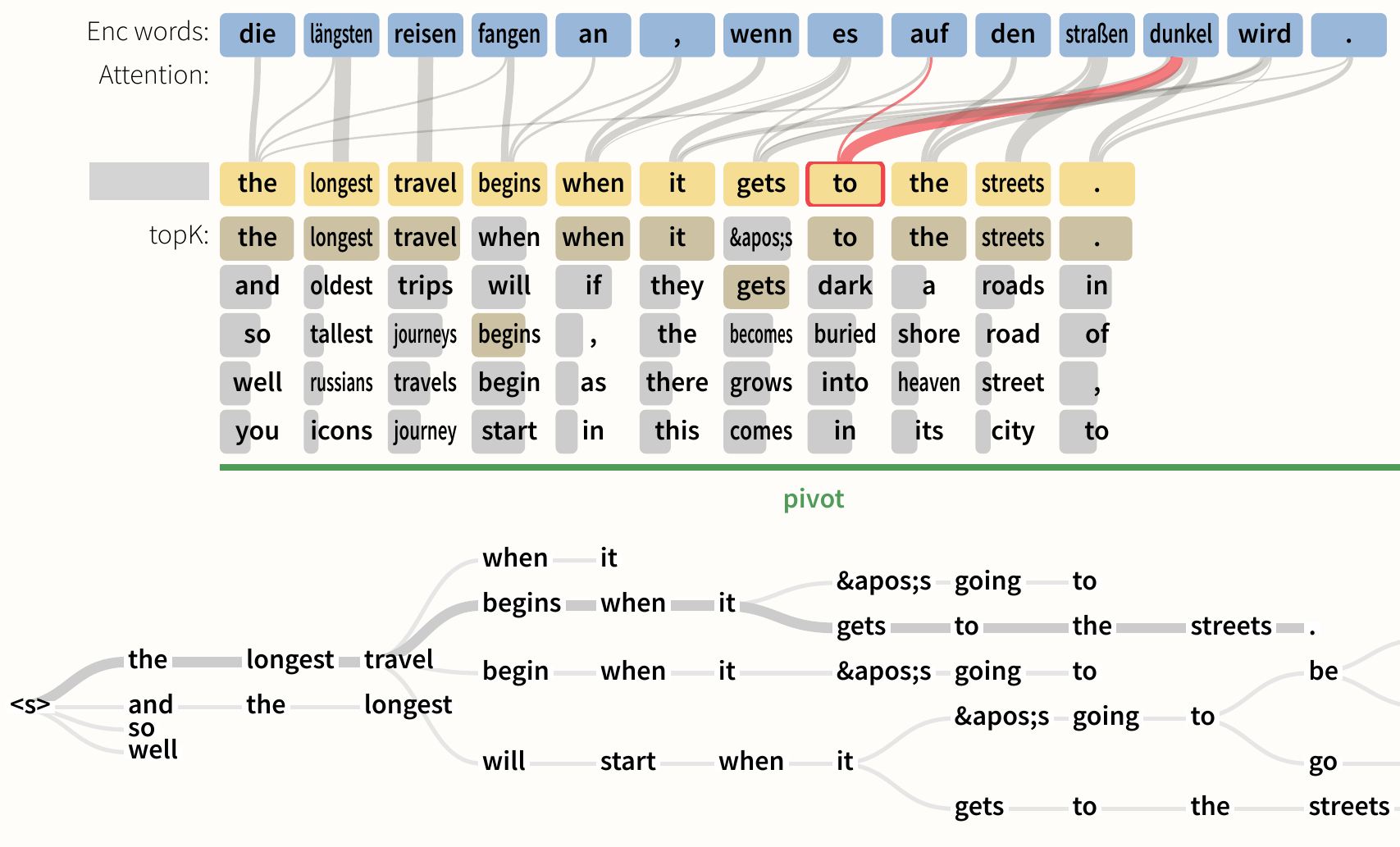}
  \caption{An example for visually debugging a sequence-to-sequence model called Seq2Seq-Vis~\cite{hen_s2s}. The highlighted attention can be excluded as likely cause for a missing context in the output sentence.  }
  \label{fig:s2s}
\end{figure}

Seq2Seq-Vis \cite{hen_s2s} is an early example of a debugging tool that helps identify which part of a sequence-to-sequence model is failing for a given instance. In this case, the encoder and decoder are LSTMs that are connected by a simple attention model. The decoder produces text using beam search. \autoref{fig:s2s} shows an example of the user interface that exposes the tokenized input sequence (blue boxes), the output sequence (yellow boxes), the attention between encoder and decoder as a bipartite graph, and the beam search tree as node-link diagram (bottom). The input sentence in German ``die laengsten reisen fangen an, wenn es auf den strassen dunkel wird'' should translate to something like \textit{the longest journeys begin when it gets dark in the streets}, but the context of \textit{dark streets} is not reproduced in the output. The red highlighted lines show that at the appropriate position in the output sentence the attention is correctly focusing on the context of \textit{dunkel} (dark). So it seems that the attention mechanism is not likely the cause of error in this case. After excluding the encoder and decoder embeddings as error-causing factors (not shown here), it seems that the beam search is not doing a good job in avoiding local optima - it prefers the slightly more likely word \textit{to} over the word \textit{dark} at the highlighted position. 

After identifying a probable cause for error, the user now can conduct what-if testing by constraining the beam tree to use the word \textit{dark} instead of the word \textit{to}. The resulting sentence \textit{The longest travel begins when it gets dark in the streets} is a very good translation. A model analyst can now add this case to a list of well-described failure examples that can later help to improve the model. 

This use case exemplifies how visual analysis of multiple parts in a complex model can help identify errors in a translation model.  While in this case, only one attention head had to be investigated, the need for more advanced methods to find and investigate relevant attention patterns is immanent in light of the rise of importance of transformer models. Visual analysis systems like BertViz~\cite{vig-2019-multiscale} or RXNmapper-VIS~\cite{schwaller2021extraction} have shown early successes in relating self-attention patterns to features in language or properties of chemical reactions encoded as SMILES strings.













\section{Research Opportunities}
\label{sec:opportunities}

We have shown early evidence that interactive visualization can play an important role to help explore and explain NLP models. 
While the existing body of work is already impressive, we think that there is potential for much more collaborative work ahead.
We base our prediction on a set of research questions that we want to highlight in this section. 

Before explaining research question related to new advances in NLP, we want to highlight a set of visualization challenges that are known to be long-standing and important to revisit in the future:
    
    \textbf{Text summarization} is a classic task in NLP and visualization. In both domains the goal is to generate abstractions/summaries for longer texts to facilitate consuming the most important information in a compressed form. One exemplary challenge is, contrary to image content, the discrete nature of text prevents the use of simple zoom out techniques for text visualization. 
    
    For language or model analysis, it is common to have multiple annotations for the same token in a text (e.g., POS tag and NER tag). Visualizing the \textbf{overlay of multiple tags for tokens} is perceptually hard and a well-known, yet unsolved, challenge in visualization.
    
    When dealing with LLMs, the amount and variety of data that is produced during inference and training challenges the \textbf{visual and interaction scalability} of any visual analytics system. Interactive visual analytic techniques for massive number of data points, such as embeddings, networks, clustering, and so on have to be optimized for large data, partial data, or progressive data updates. 

The following selection of research questions has no aspiration to be complete, but we would like to highlight some of the more recent challenges and opportunities for interactive visualization from data to inference:


\textbf{Tokenization} of the input text is a common first step for training and inference. Using sub-word tokens limits the growth of the vocabulary to feasible sizes and allows models to handle previously unseen words. However, it also raises questions about the semantic nature of tokens and what they actually represent since, e.g., a slight variation of a word can lead to completely different tokens. 

One of the main architectural ingredient in many recent LLMs that are based on the Transformer architecture is the heavy use of multiple dot product attentions across most of the layers. How can we \textbf{aggregate and visualize attention processes} in models with a rapidly increasing number of layers?

During model training or fine-tuning, checkpoints are being created that are evaluated for their task performance. But how can we quickly \textbf{compare model checkpoints} to determine qualitative progress? How can we compare models beyond highly abstract overall measures such as the computed loss? Is the increase in accuracy from $95.1\%$ to $95.11\%$ worth the training cost of our model?

A core question in XNLP is if and how we can map model-internal representations to language features. Can we identify \textbf{what the model units have learned about language}? How do we represent this knowledge? How can we make this knowledge interactively actionable? The majority of work in interactive XNLP has contributed to this topic by building tools to formulate hypotheses.

Large language models are often trained proprietorially without access to internal model states. Even with the release of weights (e.g., \citep{sanh2021multitask,meta_opt}), running the models in inference is very costly. This requires new approaches for XNLP methods. We believe that \textbf{interactive what-if testing} can play a major part in formulating hypothesis about model behavior. Related research questions are: \textit{How} can users meaningfully interact with LLMs? What are appropriate user interactions and algorithmic methods to achieve steerability in LLMs? If inference times are long, how can user interaction help to reduce the amount of LLM requests while still allowing analysts to gain insights into what the model does? 

The \textbf{limitations of language models} range from performance limits to learned biases. While model cards (\cite{Mitchell_2019}) are a good start to statically summarize how a model was trained and which biases it might expose, we think that adding interaction (beyond \cite{Crisan_2022}) to the pool of model card techniques can help to discover model limitations for real world usage. How can we communicate these complex limitations? How can we construct challenge datasets for models? How can we discover systematic errors by interaction? How can we find \textit{bad apples} and reasons why they might behave badly?

After having identified errors in a model, it might not be feasible to retrain the model again but rather apply a patch or fine-tune it very specifically. How do we \textbf{``communicate'' to a model what to fix and how}? How can we generate a generalized patch for a model? How can we observe damage being done to models while trying to fix them?

Model architectures are typically evaluated purely based on their performance on benchmark datasets. However, the extent to which we can understand a model and its decision-making process is a value in itself. Designing more \textbf{comprehensible model architectures} that are easier to debug and explore with interactive visualizations yet perform competitively would go a long way toward achieving more trustworthy and responsible AI.

\section{Conclusion}
In this position paper, we try to motivate the use of interactive visualization for XNLP by  highlighting opportunities where interactive visualization might be helpful in NLP processing workflows. We have also showed some existing examples of using visualization for XNLP. So far, the research on applying interactive visualization in XNLP is still at its early stage. To help reach its full potential, the NLP and data visualization community need to work closely together to overcome the challenges posed by siloed domain expertise. We hope that this position paper by researchers from both the NLP and visualization communities can help  encourage future interdisciplinary collaborations on this important topic.  

\section*{Acknowledgements}
 Thanks to Dagstuhl for coordination of Seminar 22191 \textit{Visual Text Analytics}. 

\bibliography{anthology,custom,own}
\bibliographystyle{acl_natbib}

\appendix

\section{Additional Use Cases}
\label{sec:appendix}

Besides the above use cases, we provide two additional cases  regarding the use of visualization for data labeling and natural language generation (NLG).

\subsection{Labeling Data}

Classifying texts is a common NLP task. For example, financial analysts have access to thousands of news sources and hundreds of thousands of blogs (e.g. Meltwater), from which news articles of interest for narrow topics must be immediately alerted to key users such as portfolio managers, traders, and quantitative analysts. Too many false positives overwhelm users, while false negatives result in missing critical market signals.

It is therefore important for the subject matter experts to create accurately labeled datasets to train a well-performing classifiers.
This entails getting an understanding of the available data set (e.g., topic distribution, level of noise) to ensure that the training data is sufficiently clean and contains enough examples of interest, it entails labeling items efficiently and effectively (manually or semi-automatically), and it also entails understanding what the model has learned and how it decides to classify individual stories to better assess the quality and generalizability of the model with the annotations made up to that point.
False labels may not only impair the performance of the trained model, but also have wide implications for society.
For instance, tweets written by African Americans have been wrongly labeled as hate speech disproportionally~\cite{sap2019risk}.

Interactive visualizations may help to find these data items to label that would improve the classifier, for instance, by inspecting borderline stories that are closer to the threshold between positive and negative stories~\cite{Heimerl2012}.
In general, two-dimensional projections of data items (e.g., with t-SNE) can help to find inaccurate labels or borderline cases~\cite{Bernard2018}.
Additional indications explaining why the model classified a certain document into a specific topic further help analysts assess whether the trained classifier has learned a plausible mapping.
There are several ways to visually explain such decisions (at least parts of it), for instance, by highlighting present or absent phrases, by visualizing what the model attended to~\cite{DeRose2021}, by highlighting active neurons on individual or aggregated items~\cite{Kahng2018}, or by depicting the evolution of hidden states in recurrent neural networks on token sequences~\cite{Strobelt2018}. For example, LIT in {\autoref{fig:explain_lit_fig}} shows a) top left, a 3D embedding of statements, color-coded by classifier result; and, b) bottom center, a salience heatmap of the selected statement.  

 These markups can also aid users interpretation of the resulting alerts. For example, the predicted score can be used as a proxy for relevance, and when presented visually, allows the user to scan lists of alerts for the most relevant stories, or, indicates model issues when irrelevant stories score highly thereby indicating model quality issues which may be due to topic drift or other causes. 

\subsection{Natural Language Generation (NLG)}
\begin{figure}[htb!]
  \includegraphics[width=\linewidth]{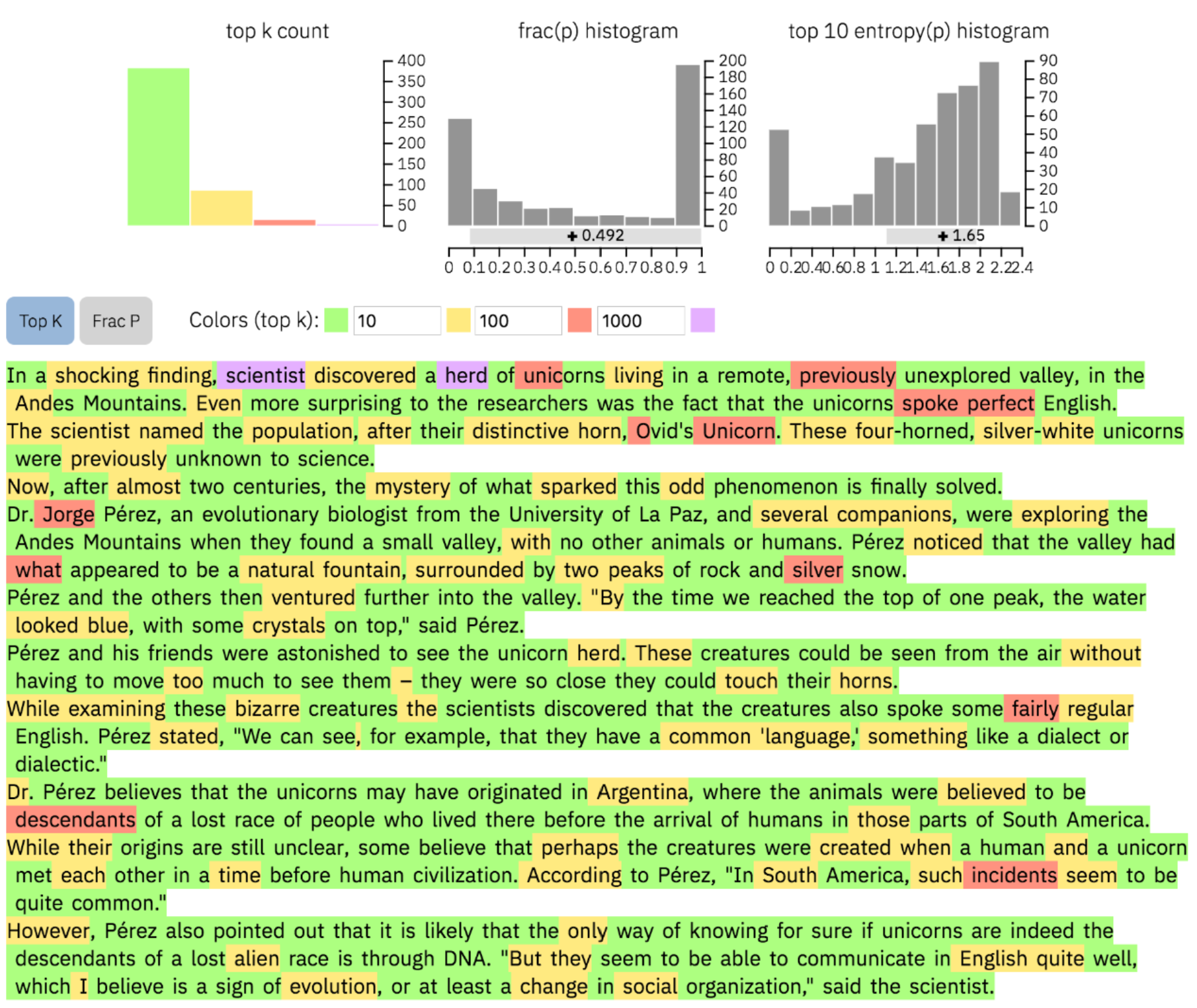}
  \caption{Example NLG markup indicating unexpected words by color. }
  \label{fig:fig_GLTR}
\end{figure}

Natural language text generation is used for tasks such as news generation, descriptive business intelligence (e.g. Narrative Science) or fiction. Text created with LLM's can result in unexpected phrases, narrative discontinuities or factual errors (e.g. hallucinations, \cite{rebuffel2022controlling}), where visualization can aid analysis and understanding of the model. GLTR~\cite{gehrmann2019gltr} colors the background of tokens based on their model probability to visualize model surprisal (e.g., red and purple indicating rather unexpected tokens with low probabilities). For instance, low overall surprisal of a given text indicates that it was either generated by the respective model or was part of its training corpus (\autoref{fig:fig_GLTR}). The interpretive color overlay on the output (a) aids text skimming to identify unexpected phrases; provides interactions, such as (b) mouse-over on the prior word shows the top subsequent words the model was expected to generate, and (c) click e.g., to regenerate from this point forward. This visualization can zoom out so that the words are no longer visible, while colored pixels remain. Markup with zoom and regeneration can be used at a macro level to skim large generated texts to facilitate editing workflows.  

Furthermore, interactions with the generated text can be used to review the training data to understand the origins of unexpected text. \autoref{fig:fig_GLTR} shows an example based on a document that was generated by another GPT-2 model. Most of the highlights are in green and yellow, indicating that the text was indeed automatically generated by a similar model. One sentence ends with \textit{two peaks of rock and silver snow} and the word \textit{silver} has a red background, corresponding to a low probability. After selecting \emph{silver snow}, a search against the training data can create a clustering visualization showing search results grouped by similarity. While \emph{silver snow} is uncommon, two large clusters indicate its use in a Nintendo game and the name of an energy drink, thus explaining to the human the origin and context of the phrase.


\end{document}